\documentclass{article}
\usepackage{spconf,amsmath,amssymb,graphicx,booktabs,balance}
\usepackage{eso-pic}
\usepackage[hidelinks]{hyperref}
\hypersetup{pdftitle={ConPro: Contrast Projection Pretraining for Label-Efficient Vessel Segmentation in DSA Sequences},
  pdfauthor={Xinge Guo, Yuanhao Wang, Liqi Shu, Yang Liu, Min Xu}}

\AddToShipoutPictureBG*{\AtPageUpperLeft{\raisebox{-12mm}{\makebox[\paperwidth]{%
\parbox{178mm}{\centering\fontsize{8}{9.5}\selectfont This work has been submitted to the IEEE for
possible publication. Copyright may be transferred without notice, after which
this version may no longer be accessible.}}}}}

\makeatletter
\long\def\@makecaption#1#2{\vskip 7pt
 \setbox\@tempboxa\hbox{#1. #2}
 \ifdim \wd\@tempboxa >\hsize #1. #2\par \else \hbox
to\hsize{\hfil\box\@tempboxa\hfil}\fi}
\makeatother

\newcommand{\Method}{ConPro}
\newcommand{\Cstar}{C^{\star}}

\title{ConPro: Contrast Projection Pretraining for Label-Efficient Vessel
       Segmentation in DSA Sequences}

\name{Xinge Guo$^{1}$, Yuanhao Wang$^{2}$, Liqi Shu$^{3}$, Yang Liu$^{2}$,
Min Xu$^{2}$\sthanks{Corresponding author: Min Xu (\href{mailto:mxu1@cs.cmu.edu}{mxu1@cs.cmu.edu}).}}
\address{$^{1}$Duke University, Durham, NC, USA \\
         $^{2}$Carnegie Mellon University, Pittsburgh, PA, USA \\
         $^{3}$University of Pittsburgh Medical Center, Pittsburgh, PA, USA}

\begin{document}
\ninept
\renewcommand{\footnotesize}{\small}%
\maketitle

\begin{abstract}
Dense vessel annotation in digital subtraction angiography (DSA) is
labor-intensive, yet every unlabeled sequence records how contrast passes
through the vessels. Semi-supervised methods take their targets from the
current model, and generic self-supervised pretexts reconstruct static
appearance, so this signal goes unused. We propose \Method{}, a
self-supervised pretraining scheme whose target is a contrast projection, the
normalized drop of every pixel below its temporal median over the sequence. On
DIAS and DSCA, with $10\%$, $20\%$ and $50\%$ of the training cases labeled,
\Method{} improves on training from scratch at every label fraction and is the
best of the compared methods on DSCA at $20\%$ and $50\%$ labels. Controlled
comparisons show that the gain comes from the target. A temporal-median target
with the same input, loss and budget stays at scratch level, and using the
projection directly instead of learning it, as an input channel or a
pseudo-label, helps little or hurts. \Method{} provides pretrained weights
without changing the segmentation architecture, so it combines with
semi-supervised training, and UniMatch, the strongest baseline, gains $0.5$ to
$2.0$ Dice and $0.9$ to $2.3$ clDice at every label fraction when started from
\Method{} weights, reaching $75.4$ Dice on DIAS and $81.3$ on DSCA.
\end{abstract}


\section{Introduction}
\label{sec:intro}

Digital subtraction angiography (DSA) is the reference examination for
intracranial stenosis, occlusion, aneurysm and moyamoya
disease~\cite{shaban2022dsa}, and it records the passage of contrast through
the cerebral vessels as a short image sequence. A mask of the arteries
supports stenosis quantification, reperfusion scoring, 3D reconstruction and
guidance during endovascular treatment~\cite{liu2024dias,zhang2025dsca}. Since
different phases opacify different branches, the mask is predicted from
several frames at once. Drawing such a mask is pixel-by-pixel work that senior neurosurgeons then
review, which is why the two public datasets contain only $60$ and $224$
labeled sequences even though DIAS alone was selected from over a thousand
acquisitions~\cite{liu2024dias}. Supervised DSA models already use the timing
of contrast arrival as an input channel~\cite{wang2024tsinet}. We use the same
signal earlier, as a mask-free pretraining target.

\begin{figure}[t]
  \centering
  \includegraphics[width=\linewidth]{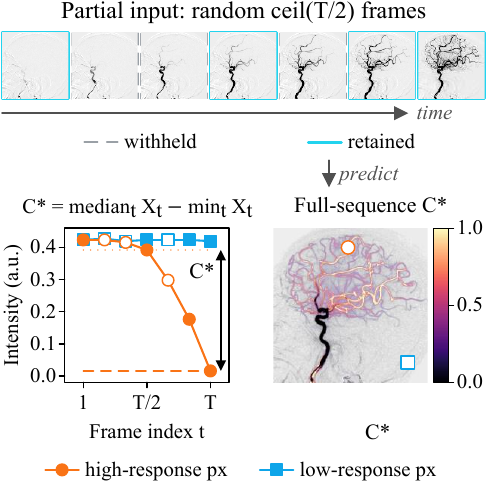}
  \caption{Annotation-free pretraining. The network sees a subset of the
  frames (cyan) and predicts the contrast projection $\Cstar$ of the whole
  sequence. Grey frames are withheld.}
  \label{fig:concept}
\end{figure}

\begin{figure*}[t]
  \centering
  \includegraphics[width=\textwidth]{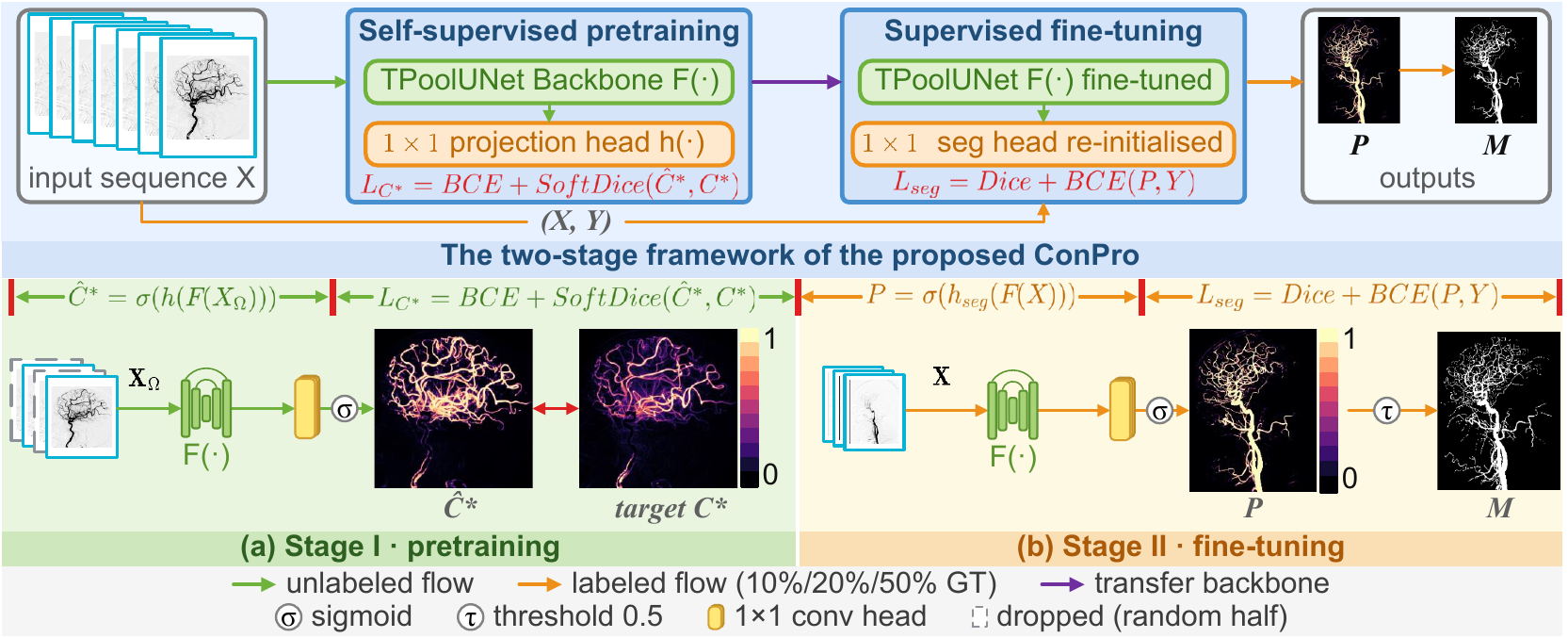}
  \caption{\Method{} pipeline. Stage~I predicts $\Cstar$ from a subset of
  frames $X_\Omega$, and Stage~II transfers the backbone $F$ to vessel
  segmentation.}
  \label{fig:pipeline}
\end{figure*}

Weak supervision replaces dense masks with cheaper
annotations~\cite{vepa2022weakdsa}. Semi-supervised methods derive
pseudo-labels or consistency targets from the current
model~\cite{chen2021cps,sun2024corrmatch,yang2023unimatch}, so their
supervision is bounded by that model. Self-supervised pretraining defines the
target from the images themselves, as masked appearance or missing
pixels~\cite{zhou2021genesis,he2022mae,tong2022videomae,feichtenhofer2022maest}.
In DSA the images carry a
target that is specific to the modality, since contrast produces a transient,
vessel-localized darkening~\cite{scalzo2016perfusion,brunozzi2018tdc}.
Angiographic pretexts to date reconstruct masked
appearance~\cite{huang2025vasomim,shao2025cmvm}, learn the subtraction
step~\cite{zeng2024pretrained}, or extract vessels
directly~\cite{ma2021ssvs,kim2023darl}, and the temporal course of the
contrast has so far been left aside as a target.

\Method{} turns this course into the supervision target
(Fig.~\ref{fig:concept}) and trains the backbone to predict it before any mask
is seen.

\noindent\textbf{Technical contributions.}
The first contribution is the contrast projection $\Cstar$, the largest
drop of every pixel below its temporal median, together with a pretraining
objective that treats it as a soft mask. The projection turns the
opacification recorded in the sequence into a supervision signal that needs
no manual annotation. To the best of our knowledge, \Method{} is the first
to use the temporal contrast projection of the DSA sequence itself, a target
tied to the vessels, for self-supervised pretraining, and to apply it to
artery segmentation of whole sequences with few labels. The second is a set of controlled
comparisons that separate the target from everything around it.
Holding input, loss and budget fixed, we swap $\Cstar$ for temporal-median
targets; holding the target fixed, we vary which frames the network sees; and
we run $\Cstar$ itself as an input channel and as a pseudo-label, with no
pretraining at all. Together these place the gain on the target and on
learning to predict it. The third is that the pretrained weights are a
drop-in initialization, so the same backbone can be handed to a
semi-supervised method, which we test with UniMatch and RPST$^{*}$.

\noindent\textbf{Medical relevance.}
\Method{} reduces the amount of annotation that sequence-level artery
segmentation needs. The pretraining runs on unlabeled sequences that any
angiography archive already holds, and the largest gain in our experiments,
$4.4$ Dice and $5.2$ clDice, is obtained with only three labeled DIAS
sequences. Crucially, the network used at inference is completely unchanged.
\Method{} alters only its initialization and adds no parameters, inputs or
computation.

\section{Method}
\label{sec:method}

Figure~\ref{fig:pipeline} summarizes \Method{}. Stage~I predicts the contrast
projection of unlabeled sequences, and Stage~II transfers the backbone to
supervised vessel segmentation.

\begin{table*}[t]
\centering
\caption{Dice\,/\,clDice (\%; three-seed means), matched backbone and Stage~II update budget.
Sup.: labeled sequences only; Semi: labeled and unlabeled sequences; Self:
self-supervised pretraining, then supervised fine-tuning. The three
$\Cstar$ rows use the contrast projection directly, without pretraining. The
last three rows are ours, \Method{} fine-tuned with labels only, and
RPST$^{*}$ and UniMatch started from \Method{} weights.
\textbf{Bold}: best; \underline{underline}: second best.}
\label{tab:main}
\vspace{2pt}
\small
\setlength{\tabcolsep}{4pt}
\begin{tabular}{llcccccc}
\toprule
& & \multicolumn{3}{c}{\textbf{DIAS}} & \multicolumn{3}{c}{\textbf{DSCA}} \\
\cmidrule(lr){3-5}\cmidrule(lr){6-8}
Method & Type & 10\% & 20\% & 50\% & 10\% & 20\% & 50\% \\
\midrule
Supervised & Sup.
 & 65.1\,/\,58.7 & 71.8\,/\,65.2 & 73.5\,/\,67.2 & 79.4\,/\,75.2 & 80.2\,/\,76.1 & 80.2\,/\,76.1 \\
CPS~\cite{chen2021cps} & Semi
 & 69.1\,/\,61.6 & 71.7\,/\,64.5 & 72.3\,/\,64.9 & 79.8\,/\,75.5 & \underline{80.5}\,/\,76.4 & \underline{80.6}\,/\,76.5 \\
CorrMatch~\cite{sun2024corrmatch} & Semi
 & 64.1\,/\,55.5 & 69.2\,/\,61.2 & 71.2\,/\,63.2 & 76.1\,/\,70.3 & 77.5\,/\,72.2 & 78.2\,/\,73.5 \\
UniMatch~\cite{yang2023unimatch} & Semi
 & 71.2\,/\,65.0 & \underline{73.1}\,/\,66.6 & \underline{74.9}\,/\,\underline{68.8} & 80.0\,/\,\underline{76.0} & 79.6\,/\,76.1 & 79.3\,/\,75.5 \\
RPST$^{*}$~\cite{liu2024dias} & Semi
 & 70.6\,/\,63.7 & 72.3\,/\,65.5 & 72.1\,/\,65.2 & 76.6\,/\,70.7 & 76.7\,/\,71.2 & 77.0\,/\,71.8 \\
Generic-SSL~\cite{zhou2021genesis,he2022mae} & Self
 & 65.7\,/\,58.6 & 70.4\,/\,64.0 & 72.2\,/\,65.8 & 78.7\,/\,74.2 & 79.9\,/\,75.9 & 80.1\,/\,76.2 \\
Supervised\,+\,$\Cstar$ channel & Sup.
 & 66.1\,/\,59.4 & 70.6\,/\,64.7 & 72.6\,/\,65.7 & 79.6\,/\,75.5 & 80.4\,/\,76.5 & \underline{80.6}\,/\,76.5 \\
$\Cstar$ pseudo-labels & Semi
 & 66.4\,/\,63.0 & 69.1\,/\,63.3 & 72.2\,/\,66.8 & 75.6\,/\,72.8 & 76.3\,/\,73.0 & 77.2\,/\,73.9 \\
$\Cstar$ channel\,+\,pseudo-labels & Semi
 & 64.2\,/\,60.4 & 65.5\,/\,61.0 & 68.2\,/\,62.3 & 75.7\,/\,72.5 & 76.2\,/\,72.9 & 75.9\,/\,72.7 \\
\textbf{\Method{}} & Self
 & 69.5\,/\,63.9 & 72.5\,/\,66.5 & 74.6\,/\,68.6 & \underline{80.2}\,/\,75.9 & \textbf{81.0}\,/\,\textbf{77.4} & \textbf{81.3}\,/\,\underline{77.6} \\
\textbf{\Method{}\,+\,RPST$^{*}$} & Self\,+\,Semi
 & \underline{71.3}\,/\,\underline{65.1} & \underline{73.1}\,/\,\underline{66.7} & 72.3\,/\,65.5 & 77.5\,/\,72.9 & 78.3\,/\,73.5 & 78.2\,/\,73.5 \\
\textbf{\Method{}\,+\,UniMatch} & Self\,+\,Semi
 & \textbf{72.5}\,/\,\textbf{66.4} & \textbf{74.5}\,/\,\textbf{68.9} & \textbf{75.4}\,/\,\textbf{69.7} & \textbf{80.7}\,/\,\textbf{76.9} & \textbf{81.0}\,/\,\underline{77.3} & \textbf{81.3}\,/\,\textbf{77.8} \\
\bottomrule
\end{tabular}
\end{table*}

\begin{figure*}[t]
  \centering
  \includegraphics[width=\textwidth]{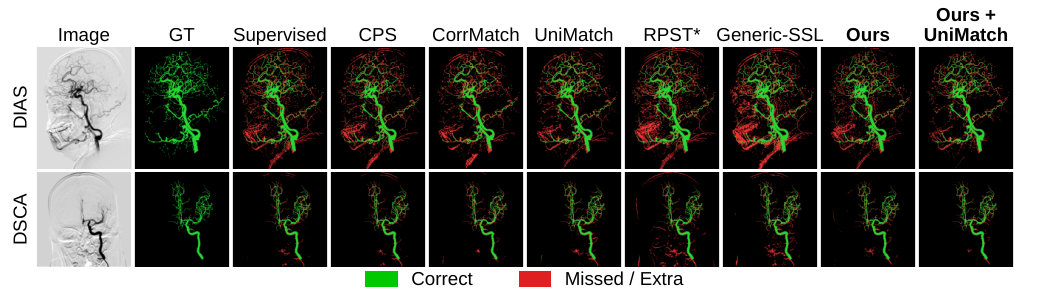}
  \caption{Qualitative results at $20\%$ labels, one test sequence per dataset.
  Green marks correct vessel pixels, red marks errors.}
  \label{fig:qual}
\end{figure*}

\subsection{Contrast projection target}
\label{ssec:proj}

Let $X=\{X_1,\dots,X_T\}$ be the valid, intensity-normalized frames of one
sequence and $x$ a pixel. In subtracted DSA, contrast transiently darkens
vessel pixels against a comparatively stable background. We encode this change
as
\begin{equation}
R(x) \;=\; \operatorname*{median}_{t} X_t(x) \;-\; \min_{t} X_t(x),
\label{eq:proj}
\end{equation}
where the median approximates the typical intensity of the pixel and the
minimum records its strongest darkening. Each sequence is then normalized to
limit extreme values:
\begin{equation}
\Cstar(x) \;=\; \operatorname{clip}\!\left(
   \frac{R(x)}{Q_{99}(R)+\epsilon},\, 0,\, 1 \right),
\qquad \epsilon = 10^{-6},
\label{eq:norm}
\end{equation}
where $Q_{99}$ is the $99$th percentile of $R$ over the sequence, computed
from the valid frames before any padding. The median reference tolerates a
first frame that already carries contrast.

\subsection{Stage~I: pretraining objective}
\label{ssec:s2f}

Let $F$ be the backbone and $h$ a $1\times1$ projection head. Given an input
set of frames $X_{\mathcal I}$, the logits $Z=h(F(X_{\mathcal I}))$ give the
prediction $\hat C^\star=\sigma(Z)$ and the loss
\begin{equation}
\mathcal{L}_{\Cstar} \;=\;
  \mathcal{L}_{\mathrm{BCE}}\!\left(Z,\,\Cstar\right)
  \;+\; \mathcal{L}_{\mathrm{Dice}}\!\left(\hat{C}^{\star},\,\Cstar\right),
\label{eq:loss}
\end{equation}
where BCE is evaluated from logits and $\mathcal{L}_{\mathrm{Dice}}$ is one
minus soft Dice~\cite{milletari2016vnet}. Both accept continuous targets.

The input set is a design choice that Sec.~\ref{ssec:abl} tests. The network
can receive all $T$ frames, or at each iteration a subset $\Omega$ drawn
uniformly with $|\Omega|=\max(2,\lceil T/2\rceil)$ and kept in temporal order.
The target can likewise be $\Cstar$ computed from all frames or
$\Cstar_\Omega$ computed from $X_\Omega$ alone. Our default feeds a random half
and keeps the full target, which amounts to frame dropout.

\subsection{Temporally pooled backbone and Stage~II transfer}
\label{ssec:backbone}

\begin{table*}[t]
\centering
\caption{Stage~I ablations (Dice\,/\,clDice \%; three-seed means). Target:
input fixed to a random half $X_\Omega$, target varied. Input: contrast
projection kept, frame feeding varied. Protocols trained independently;
\textbf{bold}: highest within a protocol.}
\label{tab:abl}
\vspace{2pt}
\small
\setlength{\tabcolsep}{1.5pt}
\begin{tabular}{@{}llcclcccccc@{}}
\toprule
\multicolumn{5}{c}{} & \multicolumn{3}{c}{\textbf{DIAS}}
& \multicolumn{3}{c}{\textbf{DSCA}} \\
\cmidrule(lr){6-8}\cmidrule(lr){9-11}
Protocol & Pretraining & Input & Target & Loss
& 10\% & 20\% & 50\% & 10\% & 20\% & 50\% \\
\midrule
Target & None & & &
 & 65.1\,/\,58.7 & 71.8\,/\,65.2 & 73.5\,/\,67.2
 & 79.4\,/\,75.2 & 80.2\,/\,76.1 & 80.2\,/\,76.1 \\
 & Median, all frames & $X_\Omega$ & $\operatorname{med}(X)$ & BCE+Dice
 & 64.1\,/\,57.6 & 70.7\,/\,63.7 & 71.7\,/\,64.3
 & 78.8\,/\,74.3 & 79.9\,/\,75.8 & 80.1\,/\,76.2 \\
 & Median, all frames & $X_\Omega$ & $\operatorname{med}(X)$ & $L_1$
 & 61.9\,/\,58.9 & 68.7\,/\,61.5 & 71.2\,/\,64.4
 & 78.8\,/\,74.3 & 79.6\,/\,75.6 & 79.9\,/\,75.9 \\
 & Median, withheld frames & $X_\Omega$ & $\operatorname{med}(X_{\bar\Omega})$ & $L_1$
 & 61.8\,/\,57.1 & 69.3\,/\,62.9 & 70.6\,/\,63.1
 & 78.7\,/\,74.3 & 79.7\,/\,75.8 & 79.7\,/\,75.7 \\
 & \textbf{Contrast projection} & $X_\Omega$ & $\Cstar$ & BCE+Dice
 & \textbf{69.3}\,/\,\textbf{63.5} & \textbf{71.9}\,/\,\textbf{66.0}
 & \textbf{74.4}\,/\,\textbf{68.4} & \textbf{80.0}\,/\,\textbf{75.8}
 & \textbf{81.2}\,/\,\textbf{77.4} & \textbf{81.4}\,/\,\textbf{77.6} \\
\midrule
Input & All frames, full target & $X$ & $\Cstar$ & BCE+Dice
 & 68.5\,/\,63.3 & 71.3\,/\,64.7 & 73.8\,/\,67.5
 & 80.0\,/\,75.7 & \textbf{81.1}\,/\,77.3 & 81.0\,/\,77.2 \\
 & Half frames, half target & $X_\Omega$ & $\Cstar_\Omega$ & BCE+Dice
 & \textbf{69.5}\,/\,63.5 & 71.9\,/\,65.8 & 74.4\,/\,68.4
 & 79.8\,/\,75.6 & 80.9\,/\,77.1 & 81.2\,/\,77.4 \\
 & \textbf{Half frames, full target} (default) & $X_\Omega$ & $\Cstar$ & BCE+Dice
 & \textbf{69.5}\,/\,\textbf{63.9} & \textbf{72.5}\,/\,\textbf{66.5}
 & \textbf{74.6}\,/\,\textbf{68.6} & \textbf{80.2}\,/\,\textbf{75.9}
 & 81.0\,/\,\textbf{77.4} & \textbf{81.3}\,/\,\textbf{77.6} \\
\bottomrule
\end{tabular}
\end{table*}

We use a lightweight temporally pooled U-Net (TPoolUNet) made of a shared 2D
encoder, masked temporal max pooling at each scale, and a U-Net
decoder~\cite{ronneberger2015unet}. For valid-frame indices $\mathcal I$, the
encoder maps each frame to features $e_t^\ell=E_\ell(X_t;\theta_E)$ at four
scales, the pool takes $\bar e_{\mathcal I}^\ell=\max_{t\in\mathcal I} e_t^\ell$
per location and channel, and the decoder $D$ produces
$F_{\theta_F}(X_{\mathcal I})=D(\bar e_{\mathcal I}^1,\ldots,\bar e_{\mathcal I}^4)$.
Restricting the maximum to $\mathcal I$ keeps padded frames out of the
pool~\cite{ng2015beyond}. Stage~I sets $\mathcal I=\Omega$ and Stage~II uses
all valid frames. The backbone has $1.08$ million parameters.

Stage~II discards $h$, initializes the backbone from the Stage~I weights,
$\theta_F^{\mathrm{II},0}=\widehat\theta_F^{\mathrm I}$, and adds a randomly
initialized $1\times1$ segmentation head $g$. For a sequence $X$ with vessel
mask $Y$, fine-tuning minimizes BCE plus Dice between $g(F_{\theta_F}(X))$ and
$Y$, as in Eq.~\eqref{eq:loss}, and the test mask thresholds the sigmoid
output at $0.5$. Labeled subsets, budget and evaluation are those of scratch
training (Sec.~\ref{sec:setup}).

\section{Experimental setup}
\label{sec:setup}

\begin{figure}[t]
  \centering
  \includegraphics[width=\linewidth]{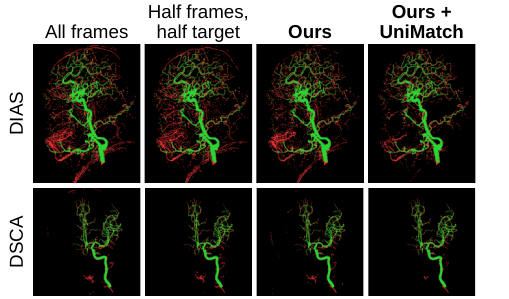}
  \caption{Input variants of Table~\ref{tab:abl} on the sequences of
  Fig.~\ref{fig:qual}, alone and with UniMatch. Colors as in Fig.~\ref{fig:qual}.}
  \label{fig:pair}
\end{figure}

\noindent\textbf{Data and evaluation.}
We use nested $10/20/50\%$ labeled subsets, $3/6/15$ of the $30$ DIAS training
cases and $15/31/76$ of the $153$ DSCA non-validation cases. DIAS further
provides $10$ validation, $20$ test and $60$ unlabeled sequences; DSCA has $27$
validation and $44$ test cases. Seeds $\{0,1,2\}$ determine subsets,
initialization and data order. Stage~I sees the $90$ DIAS training and unlabeled sequences or the $153$ DSCA
training sequences. Frames are resized to $256\times256$. DIAS sequences keep
$4$ to $14$ frames and DSCA sequences are resampled to eight. The Stage~II checkpoint is
selected by validation Dice and evaluated once on the test set. We report sequence Dice and the topology-aware
clDice~\cite{shit2021cldice}, macro-averaged over test cases.

\noindent\textbf{Compared methods.}
Every segmentation network is a TPoolUNet, and all methods share the labeled
subsets and the Stage~II update budget. CPS~\cite{chen2021cps} trains two
networks with exchanged pseudo-labels, CorrMatch~\cite{sun2024corrmatch}
propagates pseudo-labels through a correlation head, and
UniMatch~\cite{yang2023unimatch} enforces weak-to-strong consistency, all
following the official implementations. RPST$^{*}$ is our shared-backbone implementation of
RPST~\cite{liu2024dias}, one self-training round with equal numbers of labeled
and pseudo-labeled $128\times128$ patches. Generic-SSL is a
masked-reconstruction control in the style of Models Genesis and
MAE~\cite{zhou2021genesis,he2022mae} that predicts the all-frame temporal
median from a patch-masked sequence with an $L_1$ loss, everything else being
identical to \Method{}. Three controls use $\Cstar$ without any pretraining
under the Stage~II protocol, as a second input channel of every frame, as a
pseudo-label (thresholded at a value chosen on the validation set) for the
same unlabeled pool that \Method{} pretrains on, and as both at once. Since
\Method{} yields a checkpoint of the same backbone, we also run UniMatch and
RPST$^{*}$ from it.

\noindent\textbf{Ablations.}
Table~\ref{tab:abl} has two independently trained protocols. In \emph{Target}
every model receives $X_\Omega$ and only the target changes.
Here $\operatorname{med}(X)$ is the temporal median of all frames,
$\operatorname{med}(X_{\bar\Omega})$ that of the withheld frames
$\bar\Omega=\{1,\ldots,T\}\setminus\Omega$, and $L_1$ is mean absolute error.
\emph{Input} keeps the contrast projection and varies whether the network
sees all frames or a random half, and whether the target comes from all frames
or from the same half.

\noindent\textbf{Optimization and statistics.}
Stage~I runs $8{,}000$ AdamW~\cite{loshchilov2019adamw} updates (constant
rate, weight decay $10^{-4}$, batch $4$, Stage~II augmentation) and transfers
the final checkpoint as is. Stage~II runs $1{,}200$ AdamW updates (initial
rate $10^{-3}$, cosine decay), validated every $150$ updates. Supervised
fine-tuning uses four labeled sequences per batch; semi-supervised methods add
four unlabeled sequences. Pretrained encoders use $0.3\times$ the base
learning rate. Differences in Table~\ref{tab:main} were checked with paired
sign-flip tests on per-sequence scores, Holm-corrected over all comparisons.

\section{Results}
\label{sec:results}

\subsection{Performance with limited labels}

\Method{} has a higher three-seed mean than supervised training on both
metrics at every label fraction of both datasets (Table~\ref{tab:main},
Fig.~\ref{fig:qual}). The gain is largest at DIAS $10\%$, $4.4$ Dice and $5.2$ clDice, and between $0.7$
and $1.5$ points elsewhere. The Holm-corrected paired test confirms the gain
at DIAS $10\%$ and at DSCA $20\%$ and $50\%$ on both metrics and at DSCA
$10\%$ on Dice ($p\le0.01$). At DIAS $20\%$ and $50\%$ the gain is not
significant.

UniMatch is the strongest baseline. On DIAS it scores $0.2$ to $1.7$ points
above \Method{} fine-tuned with labels only, and none of these differences is
significant ($p=1.00$ after correction). On DSCA at $20\%$ and $50\%$ labels
the order reverses, and \Method{} leads UniMatch by $1.4$ to $2.0$ Dice and
$1.3$ to $2.1$ clDice ($p<0.01$). UniMatch, RPST$^{*}$ and CorrMatch all
fall below supervised training on parts of DSCA.

\Method{} provides pretrained weights without changing the segmentation
architecture, so it combines with these methods. UniMatch started from
\Method{} weights is above UniMatch from scratch at every label fraction, by
$0.5$ to $2.0$ Dice and $0.9$ to $2.3$ clDice, and the increase is
significant everywhere except DIAS $50\%$ Dice ($p=0.10$). This combination
is the best entry of the table on DIAS and at DSCA $10\%$.

\subsection{Effect of target and input}
\label{ssec:abl}

The Target protocol in Table~\ref{tab:abl} isolates the pretraining target.
With the same random half as input and the same BCE+Dice loss, predicting the
all-frame median leaves every mean within $0.1$ of scratch or below it,
whereas predicting $\Cstar$ raises it by $1.2$ to $5.2$ Dice and $1.4$ to
$5.9$ clDice over that median target. The median shows the vessels at rest,
whereas $\Cstar$ asks for the excursion below it, where the vessel is.

The Input protocol then varies how the frames are fed while the target stays
a contrast projection. The three variants differ little, and all of them
improve on scratch except all-frame training at DIAS $20\%$, which stays
$0.5$ points below it. Averaged over label fractions, the default (half
frames, full target) is $1.0$ Dice and $1.1$ clDice above training on all
frames on DIAS and $0.3$ Dice and $0.4$ clDice above the half-frame target,
and on DSCA the three are within $0.3$ points. Figure~\ref{fig:pair} agrees, and we keep frame
dropout as a default rather than claim it as a mechanism.

$\Cstar$ is label-free and already highlights vessels, so the three $\Cstar$
rows of Table~\ref{tab:main} test whether it needs to be learned at all. Fed
as an input channel, $\Cstar$ changes supervised training little, and on DIAS
at $20\%$ and $50\%$ labels it lowers the result, by up to $1.2$ Dice and
$1.5$ clDice. \Method{} stays $1.8$ to $4.5$ points above this control on
DIAS. Used as a pseudo-label, $\Cstar$ raises DIAS $10\%$ clDice from $58.7$
to $63.0$ and costs two to four points on DSCA, and adding the channel on top
lowers the DIAS results further. Thresholded directly at a
validation-chosen level, $\Cstar$ itself reaches $52.0$ and $59.3$ Dice. These results favor
using the contrast projection as a pretraining target rather than through the
direct-use strategies evaluated here.

\subsection{Scope and limitations}
\label{ssec:limits}

TPoolUNet pools frames with a maximum and $\Cstar$ is built from a median and
a minimum, so shuffling frames is an identity and the evidence above concerns
which contrast phases the network sees, not their order. The target inherits the subtraction, so motion artifacts enter
$\Cstar$ and vessels that never opacify are absent from it. Behavior at true lesions and clinical value remain to be measured.

\section{Conclusion}
\label{sec:conclusion}

A DSA sequence already describes its own vessels, because the drop of each
pixel below its temporal median marks where contrast arrived. \Method{} trains
the backbone to predict this map from unlabeled sequences and then fine-tunes
it with whatever labels exist. On DIAS and DSCA this initialization improves
on training from scratch at every label fraction, gives the best results of
the compared methods on DSCA at $20\%$ and $50\%$ labels, and also improves
UniMatch, the strongest baseline, when UniMatch starts from its weights. The
ablations put the credit on the target itself, and everything the method needs
is already in an angiography archive.

\section{Compliance with Ethical Standards}
This study was conducted retrospectively using human subject data made
available in open access by DIAS~\cite{liu2024dias} and
DSCA~\cite{zhang2025dsca}. Ethical approval was not required, as confirmed by
the dataset licenses.

\section{Acknowledgments}
No funding was received for conducting this study. The authors have no
relevant financial or nonfinancial interests to disclose.

\makeatletter\let\old@thebib\thebibliography
\def\thebibliography#1{\old@thebib{#1}\setlength{\itemsep}{0pt}}\makeatother
\bibliographystyle{IEEEbib}
\bibliography{refs}

\end{document}